\documentclass[letterpaper, 10 pt, conference]{ieeeconf}  

\IEEEoverridecommandlockouts                              
\usepackage{graphicx} 

\title{\LARGE \bf
Design of a Hybrid Robot Control System using Memristor-Model and Ant-Inspired Based Information Transfer Protocols*
}

\author{Ella Gale$^{1,2}$, Ben de Lacy Costello$^{1}$ and Andrew Adamatzky$^{1,2}$
\thanks{*This work was supported by EPSRC grant number EP/H01438/1}
\thanks{$^{1}$E. Gale, B. de Lacy Costello and A. Adamatzky are part of the Unconventional Computing Group, University of the West of England, Bristol, BS16 1QY, England
        { \{ella.gale,ben.delacycostello,andrew.adamatzky\}@uwe.ac.uk}}%
\thanks{$^{2}$E.Gale and A. Adamatzky are also affiliated with the Bristol Robotics Laboratory, Bristol, BS16 1QY, England
       }%
}

\begin{document}

\maketitle
\thispagestyle{empty}
\pagestyle{empty}

\begin{abstract}
It is not always possible for a robot to process all the information from its sensors in a timely manner and thus quick and yet valid approximations of the robot's situation are needed. Here we design hybrid control for a robot within this limit using algorithms inspired by ant worker placement behaviour and based on memristor-based non-linearity.  
\end{abstract}

\section{INTRODUCTION}

The human mind can sift through vast amounts of incoming data, disregarding currently unimportant data, yet be still able to shift attention when previously-ignored data becomes relevant. Human beings are also very good at making approximations and acting on incomplete information. Both these behaviours would be useful for robots, and to design them in requires the consideration of the bandwidth of input sensory data, the processing (thinking) time and the freshness of that data. 

Robotic control deals with how the sensing and action of a robot is coordinated. There are three timescales robot control schemes can be considered within and each has advantages and drawbacks. The shortest timescale response is governed by `reactive control', which leads to simple and predictable behaviour. The longest timescale response is `deliberative control' where the robot only acts after time spent processing all available information; this allows for more nuanced responses and has obvious drawbacks in fast-changing situations. Obviously, some combination of both extremes of control mechanisms is desirable and there have been two different solutions to this problem: `hybrid control' where part of the robot's brain plans and part deals with reactive tasks and both parts report to a controller part (and is often called a three-layer system); `behaviour-based control' which acts like the hybrid system, but in a more biological, parallel and distributed way~\cite{Mataric}. 

Memristors are novel electronic components~\cite{Strukov}, interesting as they possess a memory, are capable of learning and have low power consumption. Recent papers have suggested that neurons in the brain are memristive in action~\cite{247,248}, simulations have shown that memristors can be used as synapses within spiking neuron systems~\cite{STDP,David} and recent experimental results have demonstrating self-initialising memristor spikes~\cite{Mu0}. For these reasons it has been suggested the memristors could be used to build a neuromorphic computer. There are two main memristor theories: the phenomenological model~\cite{Strukov}, which is based on a 1-D model of variable resistors and which has been the basis of more complex models (such as those which include non-linear drift~\cite{94} or window functions) and many simulations (such as~\cite{84}); and the memory-conservation model~\cite{F0} which is based on the electrodynamics of a 3-D model of variable resistors. Because memristors are nonlinear 
components, we suggest that they could be useful in modeling nonlinear environments. 

each site in turn. The main difference between the three algorithms was the time taken to transfer the information, which was significantly different due to the nonlinear dynamics of the transfer process. This algorithm was then studied from the point of view of information transfer in a nano communication network~\cite{A1}, specifically by the situation of many sensors throughout an environment which are transmitting much more information (of different importance and quality) than the central sensing computer can deal with before the information is updated.

Previous work by our group has focused on using memristors to model the nonlinear degradation of an environment undergoing harvesting by a group of autonomous agents~\cite{A0}, this algorithm was then studied from the point of view of information transfer in a nano communication network~\cite{A1}, specifically by the situation of many sensors throughout an environment which are transmitting much more information (of different importance and quality) than the central sensing computer can deal with before the information is updated. We are interested in modeling the case where the robotic control system (RCS) has a smaller processing bandwidth than the robot's sensors can transmit to it. A recent example is the Mars rover where the robot has a great amount of data to transfer, limited bandwidth and we require that the most important (i.e. scientifically interesting) and the most mission critical (i.e. position and danger updates) data is transferred first, but we also want as much data as possible. 



In this short paper we will discuss how our previously developed algorithms can be used to take information from a robot's sensors to build up different levels of approximations cogent with the different timescales of control, what such a controller might look like and how it could be built. We won't discuss how the importance of the information is ranked, presuming that the robot designer can allocate a look-up table for that, neither will we discuss sorting the information, assuming that this can be done at the sensor or as pre-processing.

\section{APPLICATION OF ANT-INSPIRED MEMRISTOR-MODELED ALGORITHM TO A ROBOT}

The memristors were modeled using the simplest version (i.e. missing the non-linear drift) of the phenomenological model~\cite{Strukov} and, as in~\cite{A1}, this model was applied to an information source. $\beta$ is a measure of the non-linearity and is related to the decay of sensor data freshness, essentially it is a measure of diminishing returns. $R_{\mathrm{Off}}$ is fixed at the value of complete data transfer (and is the maximum resistance in~\cite{Strukov}). $R_{\mathrm{On}}$ which varies between 0 and 100 and is a measure of $\frac{1}{\mathrm{richness}}$, where richness is a convolution of bandwidth requirement and importance, perhaps as simple as the product of a number representing a measure of each. By using this modelling scheme, the current in the memristor simulation is the rate of information influx to the RCS and the total voltage drop represents the total processor bandwidth, with the voltage drop across a memristor models the amount of processor bandwidth (currently) applied to that 
sensor's data. Time dependence can be included by making $\beta$, $R_{\mathrm{On}}$ and $V$ dependent on $t$, allowing for time-dependence. 

This allowed us to design an algorithm~\cite{A0} for the scheduling problem of efficient assignment of workers to tasks and in testing such an approach, we looked at nature, specifically the Leaf cutter ant's approach in rich environments. With regards to information transfer, the main difference between the algorithms investigated is the time required to transfer the data. Here, we are envisaging `data chunks' which may come from a single source (i.e. a sensor), or virtualised data chunks such as from a combinatorial subroutine which combines highly important and time critical input from many sensors or even the output of a long-time averaging function. The Sequential algorithm starts with the `best' data chunk (lowest $R_{\mathrm{On}}$ value) and uses all the bandwidth for that and then takes data from each sensor/data chunk in decreasing `richness'. The All Sites algorithm spreads the bandwidth between all the data chunks in inverse relation to their richness value, i.e. more to poorer. The Leaf Cutter 
algorithm gives the best data chunk the entire bandwidth and then follows the All Sites algorithm for the rest. 

It was found, in tests of 2 to 75 memristors, in~\cite{A1} that if there was a low standard deviation in richness (especially if all data chunks were the same level of richness) the All Sites algorithm was the best. The Leaf Cutter worked best when there was a high standard deviation, especially when that involved there being one very good site. And above small numbers of data chunks (less than 20) the Sequential algorithm was always slowest, often by a long way. If the standard deviation of the richness is not known, then usually the Leaf Cutter was the best (around 80\% of the time, 10 uniform random trials). 

\begin{figure}[htbp]
      \centering
      \includegraphics[scale=0.73]{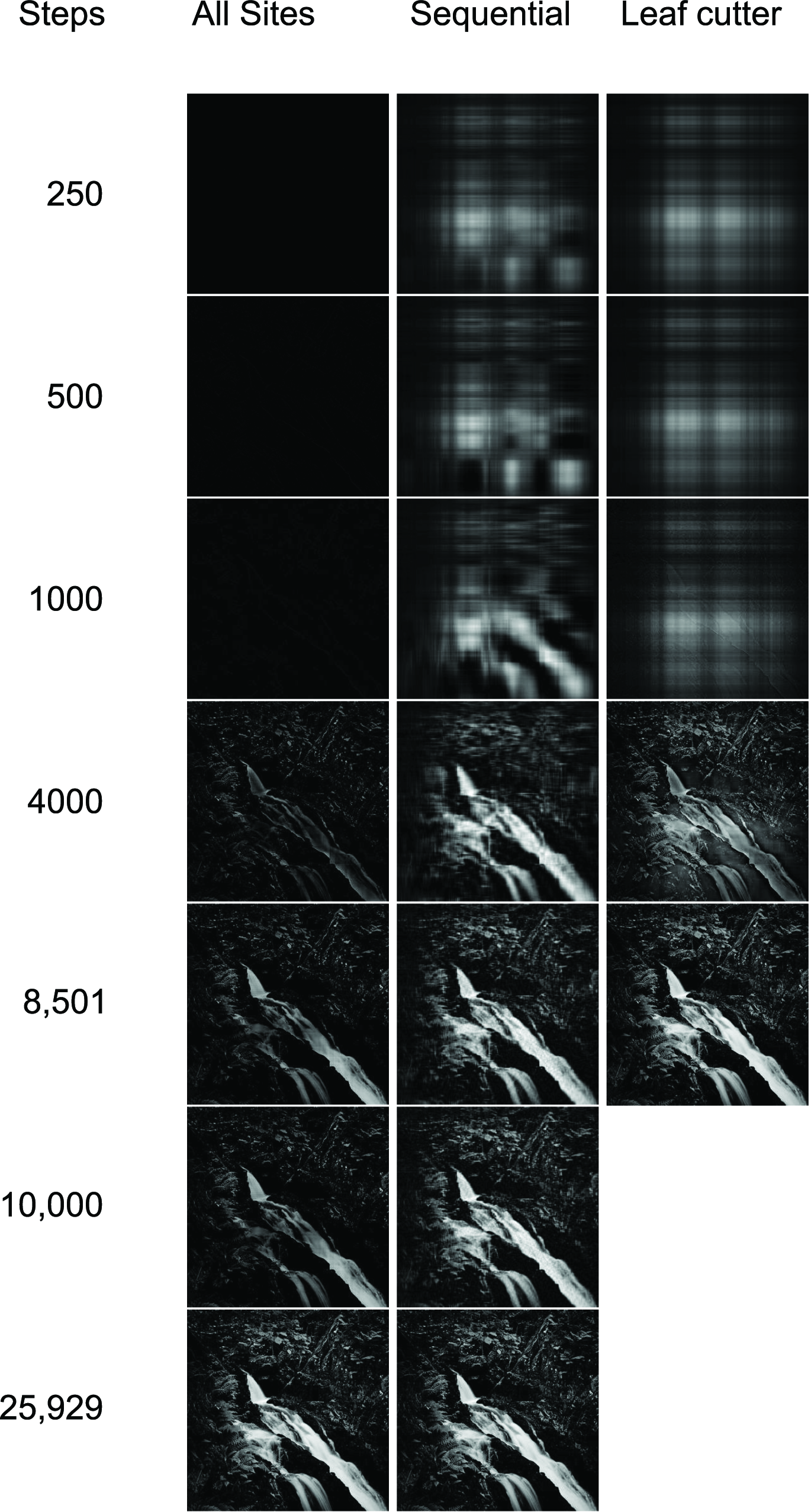}
      \caption{Information Transfer example using the singular value decomposition vectors to represent the information chunks taken from the environment via a robot's sensors. As time goes on the robot gets a more complete picture of the environment.}
      \label{fig:waterfall}
\end{figure}

More important is which of the three approaches gives the best and quickest approximation of the situation. Figure~\ref{fig:waterfall} demonstrates this for a situation where the standard deviation is relatively high as it is encoding the singular vectors of the image. If we take the final picture as the true situation and the combined transferred singular vectors as our time-dependent approximations, we can see at $< 500$ steps the Sequential algorithm has perhaps transferred the best approximation of the system. Above 500 steps the Leaf cutter algorithm approximates the environment better and takes 8501 steps to send everything compared to 25,929 for the All Sites algorithm (and 29,579 for Sequential). 

\section{APPROACHING THE CONTINUUM}

We have so far assumed that the sensors are not continually transmitting, instead, there is a call and response whereby sensors are allocated a certain time window in which to send a chunk of sampled data, and after that window, they send updated fresh data from the start again (equivalent to resetting $R_{\mathrm{On}}$). An application of this work is for approximating the environment based on continuous data transfer giving data of different freshness in the approximation. In this case, we can again refer to~\cite{A1} to see that as the number of memristors tends to infinity the Sequential algorithm scales linearly. The Leaf cutter and All Sites grow slower with memristor number, for example both under an eighth of the Sequential algorithm's time to gather data from a uniform richness distribution of 1000 data chunks (as modeled by 1000 similar memristors). The Leaf cutter algorithm will always beat the All Sites approach as long as the increase in time to process a $n+1^{\mathrm{th}}$ data chunk with 
partial bandwidth is less than the time taken to deplete the first data chunk with full bandwidth. 

\section{DECONVOLUTING RICHNESS TO ALLOW THE IMPLEMENTATION OF REACTIVE, DELIBERATIVE AND HYBRID CONTROL}

Previous work~\cite{A1} lumped several aspects of the environment together under `richness', here we will discuss the ramifications of separating them. We shall suggest taking $R_{\mathrm{On}}$ to only represent bandwidth and maintain a second list of importance of data chunks. Thus, we can use the importance list to take the most valuable data chunks, and this can be used for reactive control. Each time-step the RCS evaluates this critical data and responds straight away via a simple look-up table. Note that we can extend the Leaf cutter algorithm to allow instead the (All Sites parallelized) gathering of several data chunks of high enough importance, where the importance value is set by the designer based on time-criticality and relevance for reactive control. 

$R_{\mathrm{On}}$ would then be only a measure of the processing bandwidth requirement of the data chunk. We can use this value to process the remaining data chunks in a way that allows the creation of the best and timeliest approximation of the situation. This less critical data is still important and allows for deliberative control. This is the response mode similar to hybrid control methodology, in that it is between deliberative control and reactive control. It has a different accuracy dependent on how long the robot spends getting all the data before acting, and this value can be tweaked in an online manner by the situation itself, allowing the RCS to change the amount of `attention' it gives the surroundings before acting and varying its action rate dependent on the task. If $\eta$ is taken to be the number of timesteps the robot waits for before acting, where $1<\eta<N$, and $N$ is the total number of steps required to fully gather all information from the sensor. If $\eta = N$ we have deliberative 
control, changing $\eta$ between 1 and $N$ allows us to tune the hybrid control between fully reactive and fully deliberative. By allowing $R_{\mathrm{On}}$, the importance list and $\beta$ to be changeable (i.e. time dependent), we could manufacture a very reconfigurable system capable of complex, varied and nuanced behaviour.

It is possible to require the assessment of the long-term data and this is the mode that is required for the robot to learn (and would be the third level in a hybrid system). All received information is used to build up a long-term environmental model that would be required for an considered (intelligent) response and adaptability. Note that building up the RCS in this hierarchical manner allows the robot to override its reactive response when there is a reason to do so contained within higher level cognitive functions. This is similar to human responses for example to a broken boiler, where we might flinch from cold water but then shower ourselves due to the high regard cleanliness is held in.

By modeling attention, we make it possible for a robot to deal with an overwhelming amount of information coming in and only act on that which is relevant. By switching levels we can change the amount of importance applied to input stimuli dependent on the system. Within the importance assignment we could include a factor to compare the current input with previous values (this can be done via manipulation of virtualised data input chunks which are outputs of RCS processes) to make the robot more sensitive to changing inputs than static, similarly to biological processing. We can envisage a future application for a robot with similar capabilities and limitations to higher animals, namely a high number of sensors and a processing unit incapable of `consciously' registering all of it at once, which would require the ability to shift processing priorities dependent on the situation.

In building a system such as the RCS we can follow the `hybrid' control approach of building a reactive center, a deliberative centre and a control center to switch between the two. Memristors can be used to make distributed networks capable of adapting and learning, and as such, the real world memristors are thought to be good candidates for these types of systems. Thus, distributed parallelized memristor networks offer a way to create more biological-inspired control systems which are suited to behaviour-based control as they can be easily modified. Finally, as memristors are non-linear we expect that they will be good at dealing with information presented in a non-linear fashion as in this case.

\addtolength{\textheight}{-12cm}   






\begin{thebibliography}{99}

\bibitem{Mataric} M.J. Matari\'{c}, ``Behavior-Based Robotics as a Tool for Synthesis of Artifical Behavior and Analysis of Natural Behavior,'' Trends in Cognitive Science, vol. 2, pp. 82-87, March, 1998.

\bibitem{Strukov} D. B. Strukov, G. S. Snider, D. R. Stewart and R. S. Williams, ``The missing memristor found,'' Nature, vol. 453, pp.80-83, 2008.

\bibitem{247} Chua, L., Sbitnev, V. and Kim, H.: Hodgkin-Huxley Axon is made of Memristors, Int. J. Bifur. Chaos, vol. 22, pp.1230011(1) --1230011(48), 2012.

\bibitem{248} Chua, L., Sbitnev, V. and Kim, H.: Neurons are Poised Near the Edge of Chaos, Int. J. Bifur. Chaos, vol. 22, pp.1250098(1) --1250098(49), 2012.

\bibitem{Chua1971} L. O. Chua, ``Memristor - the missing circuit element,'' IEEE Trans. Circuit Theory, vol. 18, pp. 507-519, 1971.

\bibitem{STDP} K. Cantley, A. Subramaniam, H. Stiegler, R. Chapman, and E. Vogel, ``Hebbian Learning in Spiking Neural Networks with Nano-Crystalline Silicon TFTs and Memristive Synapses,'' IEEE Trans. Nanotech., vol. 10, 1066-1073, 2011.

\bibitem{David} D. Howard, E. Gale, L. Bull, B. de Lacy Costello and A.  Adamatzky, ``Evolution of Plastic Learning in Spiking Networks via Memristive Connections,'' IEEE Trans. Evol. Comp. vol. 16, pp. 711-729, 2012.

\bibitem{F0} E. Gale, ``The Memory Conservation Theory of Memristance,'' Unpublished, (arXiv:1106.3170v2)

\bibitem{Mu0} E. Gale, O. Matthews, B. de Lacy Costello and A. Adamatzky, ``Beyond Markov Chains, Towards Adaptive Memristor Network-based Music Generation,'' 1st AISB Symposium on Music and Unconventional Computing, Annual Convention of Society of the study of Artificial Intelligence and the Simulation of Behaviour (AISB) 2013, Exeter, UK

\bibitem{94} D. B. Strukov and R. S. Williams, ``Exponetial Ionic Drift: Fast Switching and Low Volatility of Thin-Film Memristors,'' Appl. Phys. A., vol. 94, pp. 515--519, 2009.

\bibitem{84} Y. V. Pershin and M. Di Ventra, ``Solving Mazes with Memrsitors: A Massively Parallel Approach,'' Phys. Rev. E., vol. 84, pp. 046703(1)--046703(6), 2011.

\bibitem{A0} E. M. Gale, B. de Lacy Costello and A. Adamatzky, ``Comparison of Ant-Inspired Gatherer Allocation Approaches using Memristor-Based Environmental Models,'' Bioadcom 2011 Workshop on Bio-inspired Approaches to Advanced Computing and
Communications, Bionetics 2011, York, United Kingdom 

\bibitem{A1} E. Gale, B. de Lacy Costello and A. Adamatzky, ``Memristor-based information gathering approaches, both ant-inspired and hypothetical,'' Nano Comm. Net., vol. 3, pp 203-216, December 2012


\end{thebibliography}
\end{document}